\documentclass[letterpaper, 10 pt, conference]{ieeeconf} 

\IEEEoverridecommandlockouts                             
\usepackage{booktabs}
\usepackage{amsmath}
\usepackage{amssymb}
\usepackage{hyperref}
\usepackage{graphicx}
\usepackage{subcaption}
\usepackage{xcolor}
\usepackage{tabularx}
\usepackage{dblfloatfix}
\usepackage{float}
\usepackage{multirow}
\usepackage{pifont}

\newcommand{\figref}[1]{Fig.~\ref{#1}}
\newcommand{\tableref}[1]{Table~\ref{#1}}

\title{\LARGE \bf
Procedural Generation of Articulated Simulation-Ready Assets
}

\author{
\textbf{Abhishek Joshi}$^{\dagger}$,
\textbf{Beining Han}$^{\dagger}$,
\textbf{Jack Nugent}$^{\dagger}$,
\textbf{Max Gonzalez Saez-Diez}$^{\dagger}$,
\textbf{Yiming Zuo}$^{\dagger}$, 
\textbf{Jonathan Liu}$^{\dagger}$,\\[0.5em]
\textbf{Hongyu Wen}$^{\dagger}$, 
\textbf{Stamatis Alexandropoulos}$^{\dagger}$,
\textbf{Karhan Kayan}$^{\dagger}$,
\textbf{Anna Calveri}$^{\dagger}$,
\textbf{Tao Sun}$^{\dagger,\ddagger}$,
\textbf{Gaowen Liu}$^{\S}$,\\[0.5em]
\textbf{Yi Shao}$^{\ddagger}$,
\textbf{Alexander Raistrick}$^{\dagger}$,
\textbf{Jia Deng}$^{\dagger}$\\[1em]
$^{\ddagger}$McGill University \quad
$^{\S}$Cisco \quad
$^{\dagger}$Princeton University
}

\begin{document}

\maketitle
\thispagestyle{empty}
\pagestyle{empty}

\begin{figure*}[!t]
    \centering
    \includegraphics[width=\linewidth]{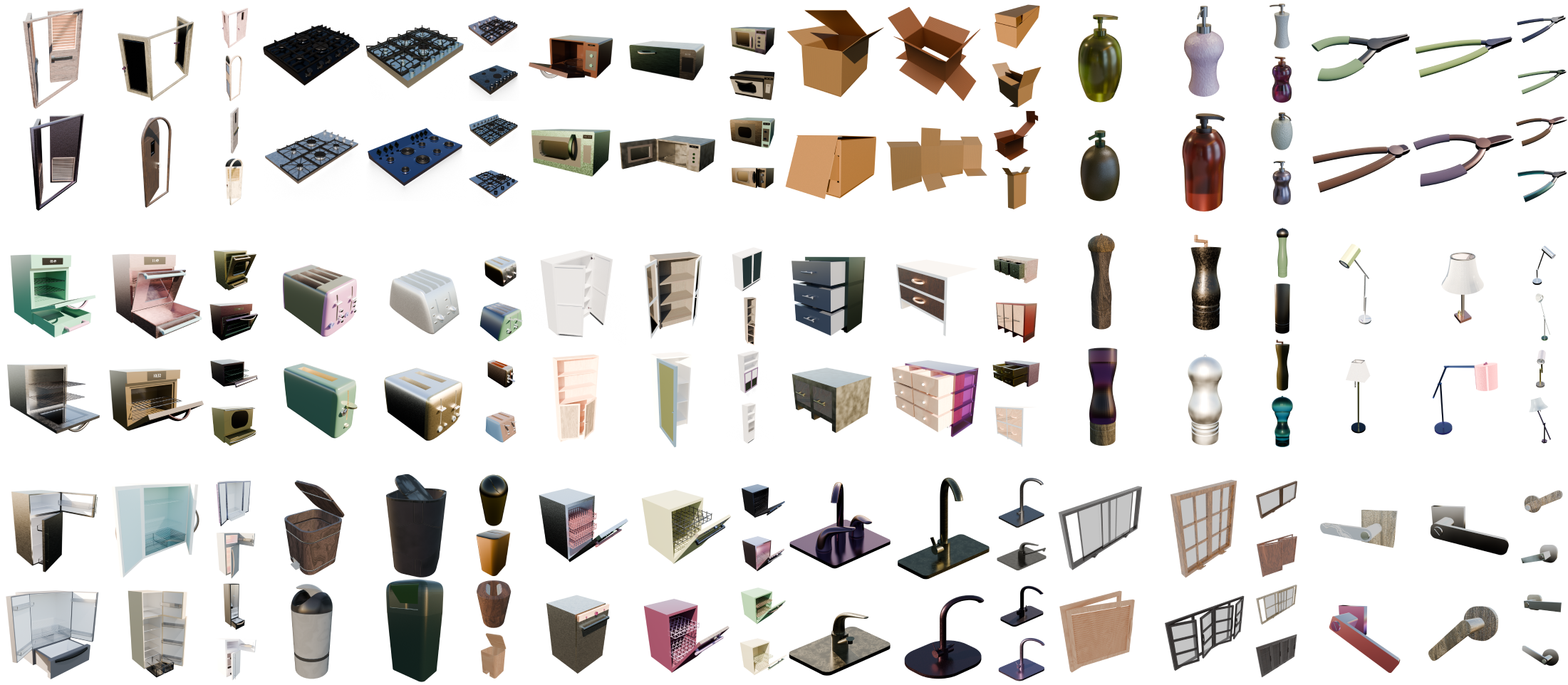}
    \caption{Procedurally generated assets for 18 common articulated object categories.}
    \label{fig:articulated_assets}
\end{figure*}

\begin{abstract}

We introduce Infinigen-Articulated, a toolkit for generating realistic, procedurally generated articulated assets for robotics simulation. We include procedural generators for 18 common articulated object categories along with high-level utilities for use creating custom articulated assets in Blender. We also provide an export pipeline to integrate the resulting assets along with their physical properties into common robotics simulators. Experiments demonstrate that assets sampled from these generators are effective for movable object segmentation, training generalizable reinforcement learning policies, and sim-to-real transfer of imitation learning policies.

\end{abstract}

\section{INTRODUCTION}

Interacting with articulated objects is an essential step in many important applications for robotics. From using hinged fridges and dishwashers, pushing buttons, opening drawers, to traversing doors with various handles, robots must manipulate these articulated objects to perform useful real-world tasks.

Learning in physical simulation \cite{taomaniskill3, robosuite2020, mittal2023orbit, procthor, li2022igibson, genesis} has proven a useful strategy for many robotics tasks \cite{quadloco2020, robocasa2024, ma2024dreureka, mujoco_playground_2025, wei2025empiricalanalysissimandrealcotraining, bjorck2025gr00t, maddukuri2025simandreal}. To best use this strategy for articulated object manipulation, we require large-scale datasets of articulated assets with as diverse geometry, joints, and visual appearance as possible. Articulated assets are challenging to acquire, as objects curated from the internet \cite{chang2015shapenet, objaverseXL} or 3D-scanning \cite{wu2023omniobject3d} are typically static and lack joint annotations. Widely used articulated object datasets instead use labor-intensive human annotation \cite{Xiang_2020_SAPIEN, liu2022abk, mao2022multiscan, behavior100, li2022behavior} for joint position and axes, which limits the quantity, quality, and diversity of unique geometries and articulation annotations.

To this end, we propose Infinigen-Articulated, a tool which enables the creation of high-quality, diverse articulated assets via procedural generation. Our tool has many advantages:

\begin{enumerate}
    \item \textit{Unlimited kinematic variation:} Procedural generators sample assets with dense coverage of task-relevant parameters that influence a robot's trajectory, e.g. the handle-to-hinge distance of a door. Continuous variation also applies to geometry detail (e.g. edge bevel radii) and tolerances (e.g. toaster slot width), which all serve to create dense coverage and diverse training data. 

    \item \textit{High-quality assets:} Joint locations, axes, and annotations can be verified for correctness unlike previous efforts which rely on human annotation or generative methods. Our assets also utilize physics-based rendering materials present in Infinigen \cite{infinigen2023infinite, infinigen2024indoors} along with material physical properties such as friction and density.

    \item \textit{Combinatorial coverage:} Procedural generators can randomly swap sub-parts to use different procedural generators (e.g. changes in handle style: knobs, levers, crash-bars). We can also easily express variable quantities of articulated parts (e.g. 1-5 fridge drawers), including very small components (e.g. tiny buttons on a toaster), which is challenging for annotators or image-based capture.

\end{enumerate}

Infinigen-Articulated builds on 3D-graphics and procedural generation tools from Blender \cite{blender} and Infinigen Indoors \cite{infinigen2024indoors} by adding tools and generators designed for articulated object generation. Specifically, we augment Blender's Geometry Nodes system with new modules (known as \textit{node-groups}) which implement hinged and sliding joints for procedural assets. These node-groups ensure that joints are represented in a unified format for use by downstream tools. We provide export code which converts assets created using these tools into high quality assets for simulation in popular formats such as URDF, USD, and MJCF.

Experiments show that objects from procedural generators created with Infinigen-Articulated improve upon baselines in multiple settings for vision and robot learning. We use Infinigen-Articulated to create articulated procedural generators for 18 object categories. We hope Infinigen-Articulated will provide an ever-growing library of open-source articulated object generators as to enable more robust robot learning across many tasks using articulated objects.

\section{Related Work}
\label{sec:result}

\textbf{Asset Datasets for Robot Simulation.} Although works have collected large-scale static 3D asset datasets \cite{objaverseXL, chang2015shapenet}, interactive and articulated 3D datasets \cite{Xiang_2020_SAPIEN, Mo_2019_CVPR, geng2022gapartnet, Liu_2022_CVPR, li2022behavior} remain limited in both quantity and quality. Sapien \cite{Xiang_2020_SAPIEN} collected 2k+ assets from PartNet \cite{Mo_2019_CVPR} and manually annotated the articulated joints. Others \cite{liu2022abk, mao2022multiscan, kim2024parahome} scanned real-world objects to curate separate articulated object datasets. However, issues like inaccurate joint annotations and a lack of kinematic diversity remain. Infinigen-Articulated instead adopts a new approach (procedural generation) which has precise ground truth and is highly scalable, only requiring a new procedural generator for each object \textit{class}, with no human time cost for each unique object geometry or articulation configuration.

\textbf{Articulated Object Generation.} Recent work has exploited AI techniques to augment and diversify static asset datasets \cite{objaverseXL, Calli_2015} to create articulated objects \cite{le2024articulate, qiu2025articulate}. However, these systems lack controllability of asset geometries and do not guarantee correct joint configurations \cite{liu2025surveymodelinghumanmadearticulated}. Prior work \cite{jiang2022ditto, le2024articulate, mandi2024real2code, liu2023paris, chen2024urdformer, Hsu2023DittoITH, qiu2025articulate, weng2024neural} has also generated articulated assets with real-to-sim pipelines. Paris \cite{liu2023paris} learns part-based neural radiance fields to reconstruct articulated assets. Real2Code \cite{mandi2024real2code} and URDFormer \cite{chen2024urdformer} constructs URDFs with RGB images as input, using a mixture of off-the-shelf vision models and LLMs. That being said, real-to-sim approaches often fail to construct articulated assets with high-quality. Additionally, these methods fall short on capturing occluded articulations and tend to produce inaccurate geometries and joint annotations \cite{liu2025surveymodelinghumanmadearticulated}. Inherited from Infinigen-Indoors \cite{infinigen2024indoors}, our procedural generators model each individual part from scratch including their joint properties. This means users can also easily create high quality articulated bodies that may be occluded by outer surfaces (e.g. racks in a dishwasher or drawers in a refrigerator).

\textbf{Procedural Generation.}  Many prior works create procedural training data focused on indoor environments \cite{procthor, infinigen2023infinite}, nature \cite{infinigen2023infinite}, and cities \cite{proccity} but do not create procedural articulated objects. More similar to our work is \cite{Eppner2024}, which introduces a scene generator including some procedural articulated assets made in Python. However, their assets are composed mainly of mesh primitives (cubes/cylinders), do not contain multi-level articulations (e.g. pivoting handle on a pivoting door), and is extensible only by programmers. Our system is constructed using Blender, has detailed geometry and materials, and is extensible by both programmers and anyone familiar with Blender.

\section{Infinigen-Articulated}

\begin{figure*}[!t]
    \centering
    \includegraphics[width=0.85\linewidth]{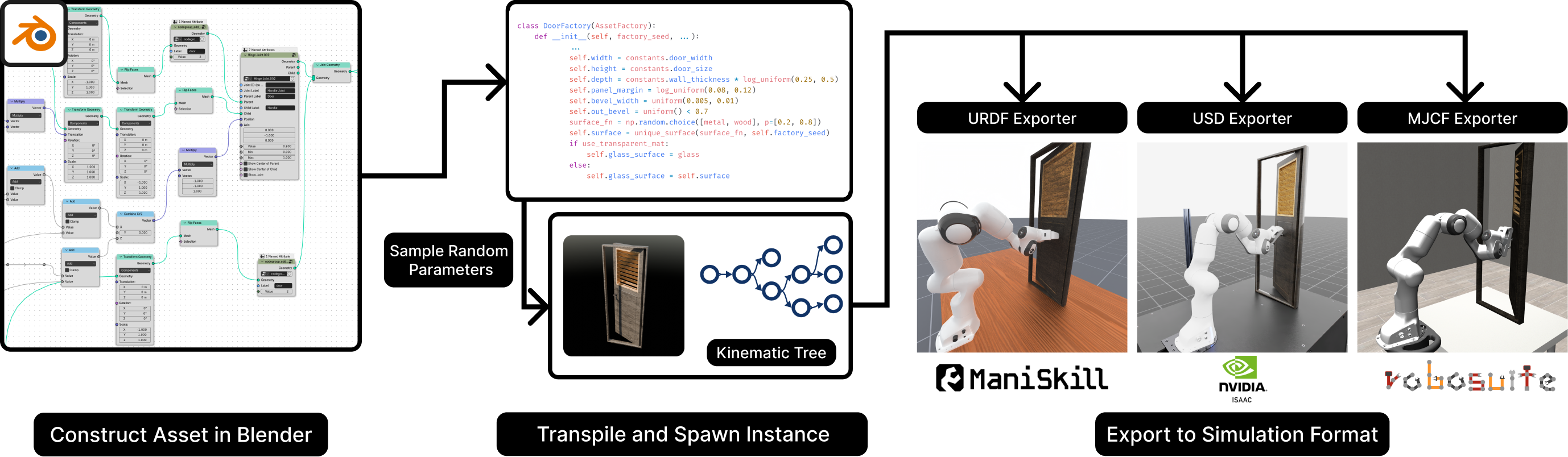}
    \caption{We first create procedural assets using Blender's Geometry Nodes feature. These assets are then transpiled, and we have the ability to define custom distributions for the procedural parameters (including materials) along with its dynamics properties. We sample parameters to generate the asset and pass both the asset and its kinematic tree to the exporter, which recursively builds a simulation-ready articulated object.}
    \label{fig:pipeline}
\end{figure*}

Infinigen-Articulated consists of a variety of procedural generators for common articulated object categories along with utilities designed to aid in creating custom generators. We also develop export code which enables articulated objects to be used in robotics simulators. The overall workflow is visualized in \figref{fig:pipeline}.

\subsection{Articulated Procedural Generator Tools}

Infinigen-Articulated leverages Blender’s procedural modeling tools \cite{blender}, namely \textit{Geometry Nodes} and \textit{Shader Nodes}, which are an artist-friendly, GUI-based, domain specific language for procedural generation. These tools allow users to create meshes and materials by composing nodes representing primitives, geometric transformations, and scalar vector arithmetic in a directed acyclic graph. These node-based tools are general purpose and can represent essentially all desirable object geometries. 

To design articulated assets within this system, we create two new nodes that implement revolute and prismatic joints. These nodes accept two incoming geometries (defined via prior nodes) and introduce a parent-child relationship annotated with the relevant joint type. Each joint node also allows users to set the pivot position, axis, and joint range, either as fixed values or through parameters passed from other nodes. For the provided generators, we leverage Blender’s built-in math nodes to compute these values directly from the asset’s geometry within the node graph.

These joint nodes have four purposes. First, they speed up the procedural generator creation process since they implement a commonly used operation which no longer needs to be re-implemented for each object that uses a hinge or sliding joint. Second, they provide immediate visual feedback as to the effect of a joint: a hinge joint node with a certain origin or angle extent input will immediately transform the child geometry in the GUI according to those inputs, which prevents erroneous values and can be used to debug any arithmetic expressions responsible for calculating the joint parameters. Third, this standardized implementation of joints allows us to add logic to save all joint metadata in a standard format for later use. Fourth, these nodes can store semantic information, ensuring consistent, fine-grained joint and geometry naming across all procedurally generated assets within a class.

Our joint nodes support the creation of many articulated structure types such as jointing rigid bodies in a sequence, jointing multiple bodies to a single parent, and adding multiple degrees of freedom between two bodies. We also create an additional custom node group that supports duplicating jointed bodies at defined points. This can be effective for procedurally duplicating articulations such as burners on a cooktop or shelves in a dishwasher without having to manually define each joint. Finally, we provide a utility script that checks if any two rigid parts of the asset penetrate each other within a defined range of motion. This helps users designing custom articulated assets adjust joint configurations and part geometries to prevent self-collisions and penetration issues.

\subsection{Exporting Assets to Simulation}

Converting instances of procedurally articulated assets from Blender to a simulation-ready format involves three steps, all of which are automated. First, we use Infinigen's \cite{infinigen2023infinite} transpiler to convert the procedural asset's node graph to Python code. Users can edit this code to customize asset parameter distributions or further tune node graph structures. Thus, our tool gives users fine-grained control over generated assets. It also allows users to create dynamic graphs that may change based on the procedural parameters.

Second, we sample random values for each procedural parameter and spawn an instance of the asset, along with its node graph, in Blender. We then use the Blender Python API to parse and update the asset's node graph. During this step, we inject into the graph Blender-provided nodes which assign values for attributes to different parts of the geometry. These attributes enable us to construct a kinematic tree for the asset instance (inspired by \cite{nap, liu2024cage}) that abstracts away the full procedural graph while preserving the essential details required for articulation.

The final step is to export the asset into a simulation-compatible format. We pass the asset instance along with its corresponding kinematic tree to a native exporter for either one of the URDF, USD, and MJCF file formats. To construct the asset, the exporter recursively traverses the kinematic tree depth-first. To get individual rigid bodies, the exporter queries the asset instance with attributes and their values based on the current path in the tree. We then extract the part of the asset that matches this query and save it as an individual mesh. Users can also generate convex-decomposed collision meshes with third-party tools like CoACD \cite{wei2022coacd}. During this process, we use the kinematic tree to build the final simulation-ready asset including all joints, rigid geometries, and semantic metadata.

\section{Articulated Asset Procedural Generators}
\label{sec:examples}

Using Infinigen-Articulated, we create procedural generators for 18 categories of common articulated assets as shown in \figref{fig:articulated_assets}. Each procedural generator includes distributions for joint dynamics parameters (i.e. stiffness and damping) and produces kinematically diverse, photorealistic articulated assets. A list of articulated parts is provided in \ref{sec:articulated_parts}.

\begin{figure}[t]
    \centering
    \includegraphics[width=\linewidth]{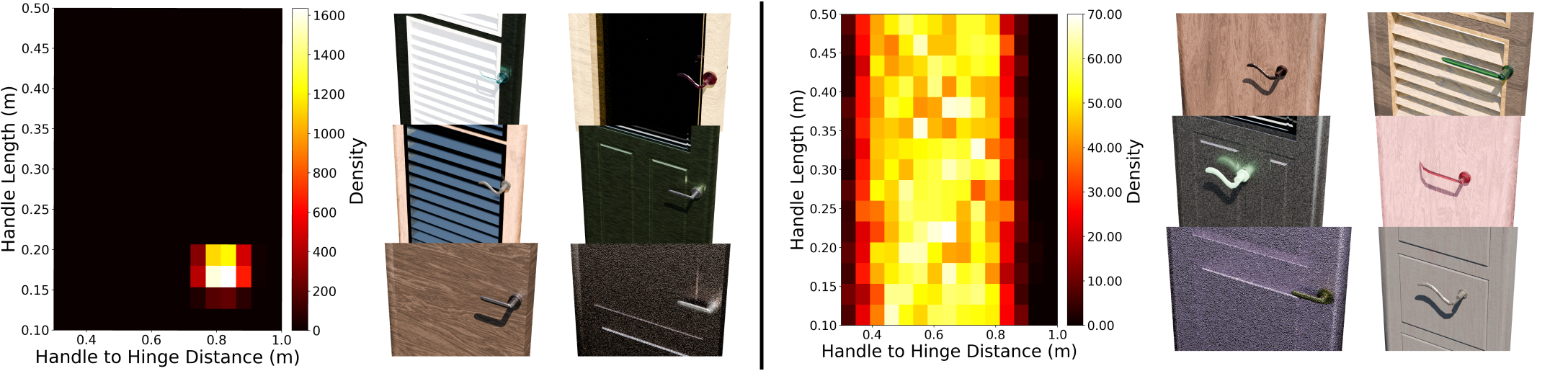}
    \caption{Comparison of asset parameter distributions. The left column shows the default distribution used for handle length and handle-to-hinge distance. The right column shows an expanded distribution where these parameters are varied over a wider range.}
    \label{fig:custom_dists}
\end{figure}

All assets created in this way have many continuously varying input parameters, which can either be manually set or randomized according to a distribution. We show a case study for doors with lever handles in \figref{fig:custom_dists}. To open a door with a handle, two parameters are critical for manipulation: the length of the handle and the distance between the handle hinge and the door hinge. The former defines grasp poses and the motion of handle rotation. The latter determines the trajectory required to successfully rotate the door. Infinigen-Articulated allows users to define widely distributed combinations of these two parameters with diverse handle geometries. Users can tailor distributions of parameters to their specific needs. Such properties cannot be satisfied with existing articulated assets \cite{Xiang_2020_SAPIEN, liu2022abk} or modeling tools \cite{le2024articulate, mandi2024real2code}.

\begin{figure}[t]
    \centering
    \includegraphics[width=\linewidth]{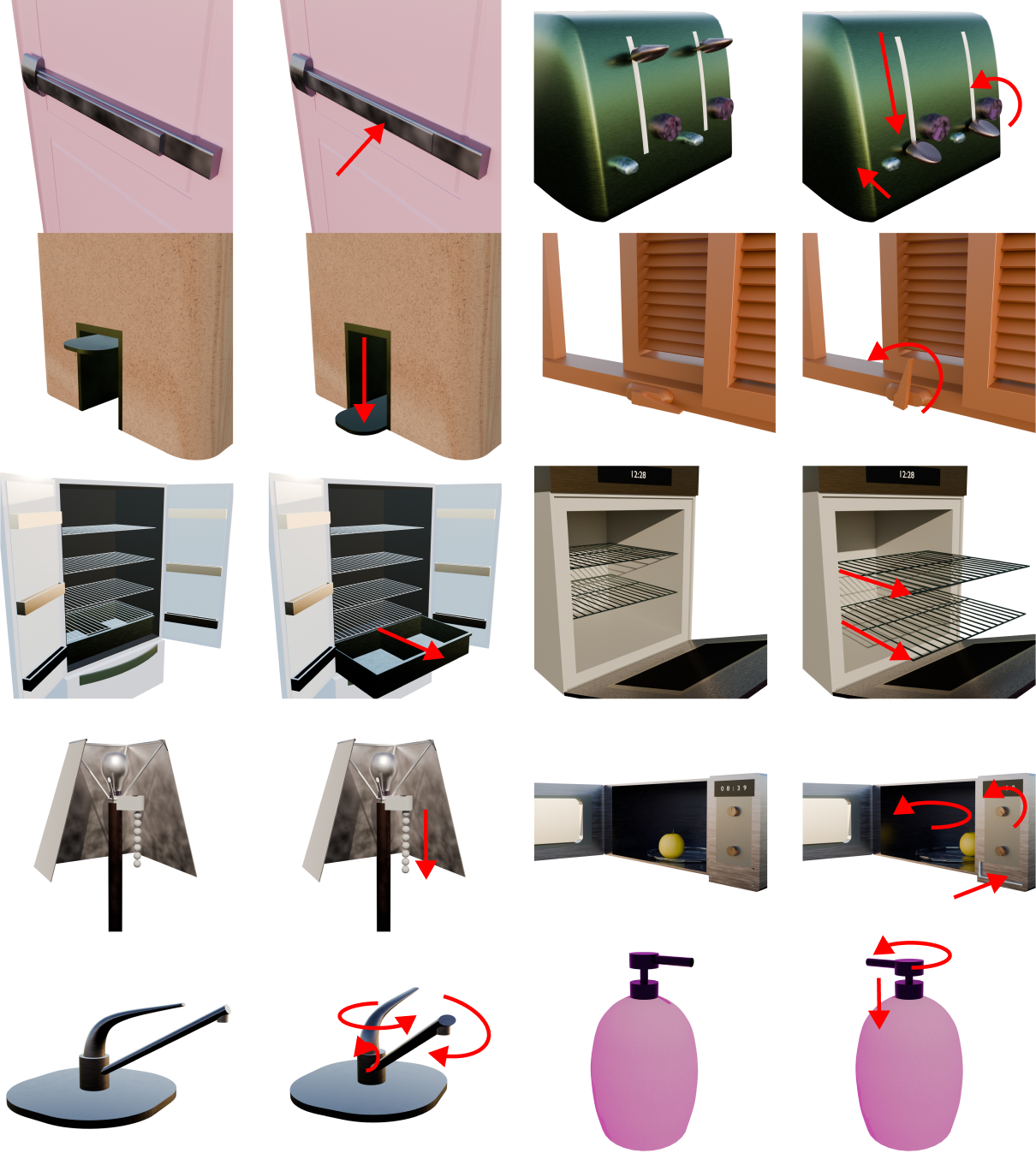}
    \caption{Examples of articulations for small, but essential parts (rows 1 and 2), occluded objects (rows 3 and 4), and double-jointed objects (row 5).}
    \label{fig:detailed_articulations}
\end{figure}

Another key advantage of procedurally generating articulated assets is the ability to scale small, detailed components. These parts are particularly important to produce robust policies for tasks such as opening a door with a lock or pulling a thin, potentially occluded, chain to turn on a lamp. Our provided procedural generators include several detailed parts, each with their own procedural parameters. This provides users the ability to scale assets with high levels of detail and accuracy (i.e. correct joint position, axis, and range) while incurring little overhead. Examples of such articulated parts can be seen in \figref{fig:detailed_articulations}. In contrast, manually annotating and articulating detailed parts of an asset is time-consuming and is much harder to scale accurately.

\begin{table*}[t]
\centering
\begin{tabularx}{0.9\linewidth}{l | *{3}{X} | *{4}{X} | *{2}{X}}
\toprule
& \multicolumn{3}{|c|}{\textbf{Door}} & \multicolumn{4}{|c|}{\textbf{Toaster}} & \multicolumn{2}{|c}{\textbf{Refrigerator}} \\
\midrule
\textbf{Dataset} & Frame & Body & Handle & Body & Lever & Knob & Button & Body & Door \\
\texttt{P15k} &  57.41  &  71.87   &   41.08  &  96.96  &  59.76  &  55.31  &   4.20   &   81.35  &  66.37  \\
\texttt{P30k}  &  59.21   &  \textbf{74.65}  &   40.31   &  \textbf{97.15}  &  59.05  &  \textbf{57.50}   &   3.70   &  82.76   &  \textbf{69.00}   \\
\texttt{P15k+I}  &  \textbf{59.64}   &  73.17  &  \textbf{44.97}   & 96.87   &  \textbf{64.31}   &  54.69   &  \textbf{5.82}  &  \textbf{84.37}   &  67.82   \\
\end{tabularx}

\begin{tabularx}{0.9\linewidth}{l | *{5}{X} | *{4}{X} | *{1}{X}}
\toprule
& \multicolumn{5}{|c|}{\textbf{Dishwasher}} & \multicolumn{4}{|c}{\textbf{Lamp}} & \multicolumn{1}{|c}{\textbf{mAP}} \\
\midrule
\textbf{Dataset} & Body & Door & Shelf & Button & Knob & Base & Rod & Head & Switch & Overall \\
\texttt{P15K}   &   84.23   &   73.32   &  0.13  &    24.59   &   0.00   &   54.75   &   \textbf{7.27}  &   \textbf{78.09}   &    11.45  &  48.23 \\
\texttt{P30k}   &    85.14  &   75.14   &   \textbf{0.21}   &    21.31   &   0.00   &   51.72   &   7.22  &   77.86   &    10.45  & 48.46  \\
\texttt{P15k+I}  &  \textbf{86.59}    &   \textbf{75.31}   &   0.17    &   \textbf{31.01}    &   \textbf{0.02}   &  \textbf{55.40}    &  7.03  &   77.63  & \textbf{17.44}  & \textbf{50.13}   \\
\bottomrule
\end{tabularx}

\caption{Movable part segmentation performance per articulated part type. We train a model for 50 epochs and report the evaluation mAP scores averaged across 3 seeds. Using Infinigen-Articulated assets results in stronger model generalization for relatively smaller articulated parts, e.g. door handles, toaster levers, dishwasher buttons, and lamp switches.}
\label{tab:map_indiv}
\end{table*}

\section{Experiments}
\label{sec:experiment}

We demonstrate Infinigen-Articulated's potential on downstream vision and robot learning tasks, including movable part segmentation, RL generalization, and sim-to-real transfer.

\subsection{Movable Part Segmentation}
\label{sec:part_seg}

Movable part segmentation \cite{moveableparts, Xiang_2020_SAPIEN} is an important vision task for embodied agents. Given an RGB image of an object, the model must segment each articulated part. We focus on five categories of assets including doors, toasters, refrigerators\footnote{The refrigerators used in the movable part segmentation experiments have since been updated (\figref{fig:articulated_assets}) with more diverse handle types, more realistic materials, and fewer vertices.}, dishwashers, and lamps. First, we generate 250 diverse assets per category using Infingen-Sim. Then, we follow a similar setup as \cite{Xiang_2020_SAPIEN} to generate image datasets for both PartNet-Mobility and Infinigen-Articulated assets. We split the PartNet dataset such that images rendered from 75\% of the assets are used for training and the remaining 25\% are used for evaluation. All models are evaluated on images of \textit{unseen} PartNet assets only. We finetune a pretrained Mask R-CNN model with a ResNet-50-FPN backbone, as in \cite{li2021benchmarkingdetectiontransferlearning}, for three datasets: 1) \texttt{P15k}, approximately 15k images of PartNet assets only, 2) \texttt{P30k}, approximately 30k images of PartNet assets only, and 3) \texttt{P15k+I}, a combination of approximately 15k images of PartNet assets and 15k images of Infinigen-Articulated assets. 

As shown in \tableref{tab:map_indiv}, we observe that doubling the size of the original dataset (\texttt{P30k}) yields only marginal improvements, and in some cases decreases performance. This suggests a saturation in the model's ability to learn new features that improve generalization to unseen assets despite scaling the training images from the original PartNet dataset alone. In contrast, we see a significant improvement when co-training with images of Infinigen-Articulated assets across smaller articulated parts such as door handles, toaster levers, dishwasher buttons, and lamp switches. This highlights a core benefit of Infinigen-Articulated. Namely, our tool allows users to procedurally scale articulated parts that require a greater level of detail. Achieving this level of granularity is labor-intensive for manually annotated articulation datasets. Despite our assets being out of distribution compared to the evaluation set, we still see improvements for detailed parts.

\subsection{RL Generalization}

We demonstrate that Infinigen-Articulated assets help reinforcement learning policies generalize to novel object instances. For each task, we train three policies, namely one trained with only Infinigen-Articulated assets, one trained with only Partnet-Mobility assets, and one trained on a combination of both. We evaluate on the following tasks:

\begin{enumerate}[leftmargin=0.2in, topsep=0pt, itemsep=0pt, parsep=0pt]
    \item \textbf{Push Door with Handle}. The robot is required to rotate the handle clockwise by 0.3 radians and push the door open. We limit the scope to single, push doors with a lever handle on the left.
    \item \textbf{Push Down Toaster Lever}. The robot is tasked with pushing down the lever of the toaster. Only toasters with a single lever are considered for this task.
    \item \textbf{Open Fridge Door}. The robot is expected to grasp the handle and open the fridge door by at least 9 degrees ($\pi/20$ radians). Only refrigerators with a single door with a handle on the left are considered for this task.
\end{enumerate}

To ensure fairness, our evaluation sets consists of a mixture of 5 \textit{unseen} Infinigen-Articulated assets and 5 \textit{unseen} PartNet-Mobility assets. We train PPO \cite{schulman2017ppo} policies in ManiSkill3 \cite{taomaniskill3} using GPU parallelized heterogeneous simulation across 128 environments, where each environment may contain a different asset. The PPO policy takes as input its end-effector pose and an RGB image rendered by the default ManiSkill3 renderer from a third-person view with no additional privileged information. The output at each time step is a delta pose command for operational space control (OSC). We refer the reader to the appendix for details.

\begin{figure}[h]
    \centering
    \includegraphics[width=\linewidth]{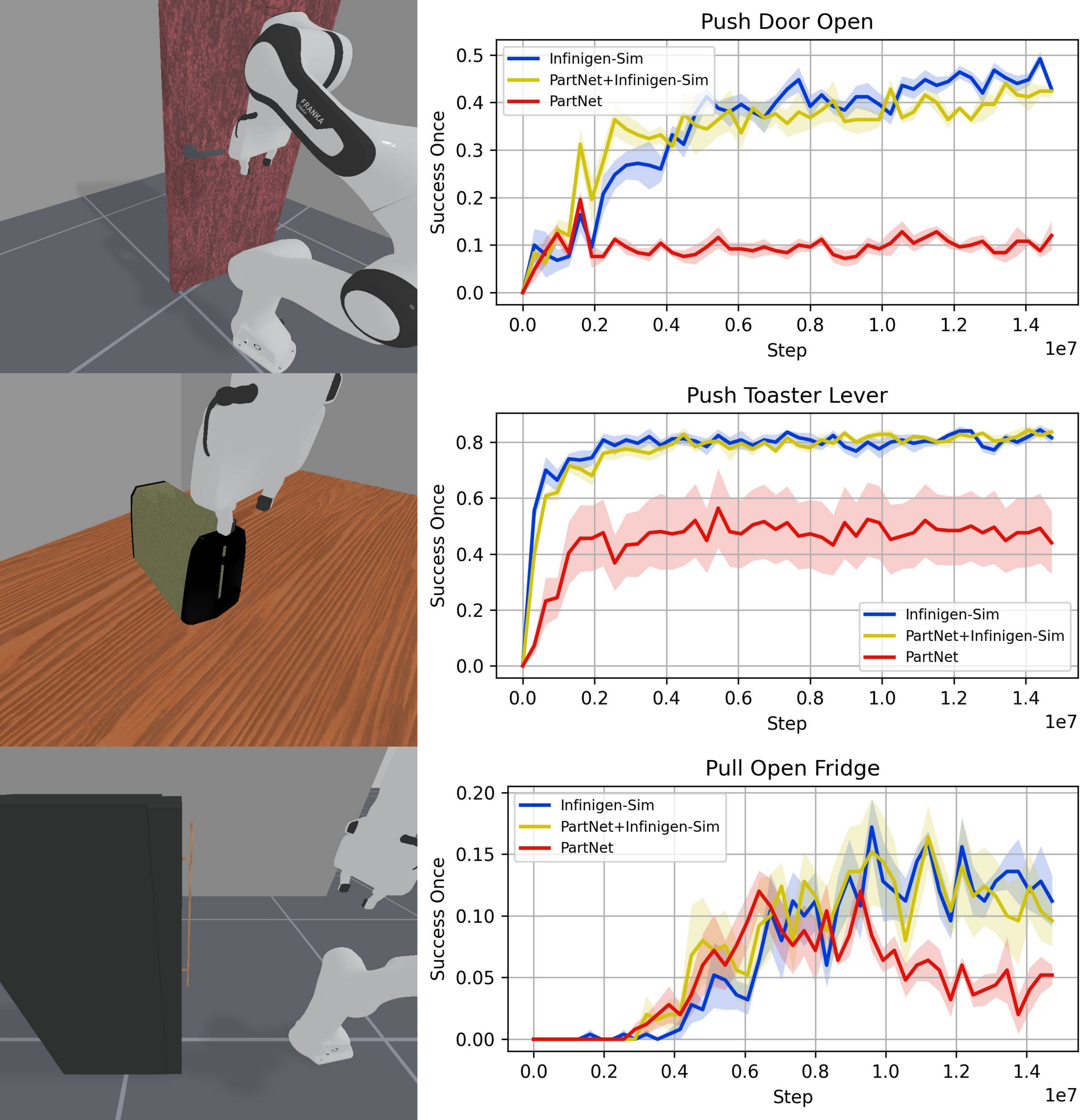}
    \caption{Left: Environment setups for each task. Right: Average success-once performance over 5 seeds with bounds showing $\pm0.5$ standard deviation.}
    \label{fig:rl_results}
\end{figure}

Our results demonstrate a 2.86-fold average improvement in success rate across all three tasks (during the last 1 million timesteps) when using exclusively Infinigen-Articulated assets over PartNet-Mobility. We attribute this performance increase to the scale, quality, and diversity of Infinigen-Articulated assets. We hypothesize that the limited number of PartNet-Mobility assets, together with potentially misaligned joints and poor geometries, significantly increases the difficulty of training effective policies. In contrast, our tool provides dynamically accurate assets with both great visual and geometric diversity. However, it is worth noting that during the earlier stages of training, for the door and fridge opening tasks, the PartNet-Mobility policy performed on par or better than policies trained using Infinigen-Articulated. This suggests that realizing the benefits of more diverse assets may require longer training times.

\subsection{Sim-to-Real Transfer}

\begin{figure}[h]
    \centering
    \includegraphics[width=\linewidth]{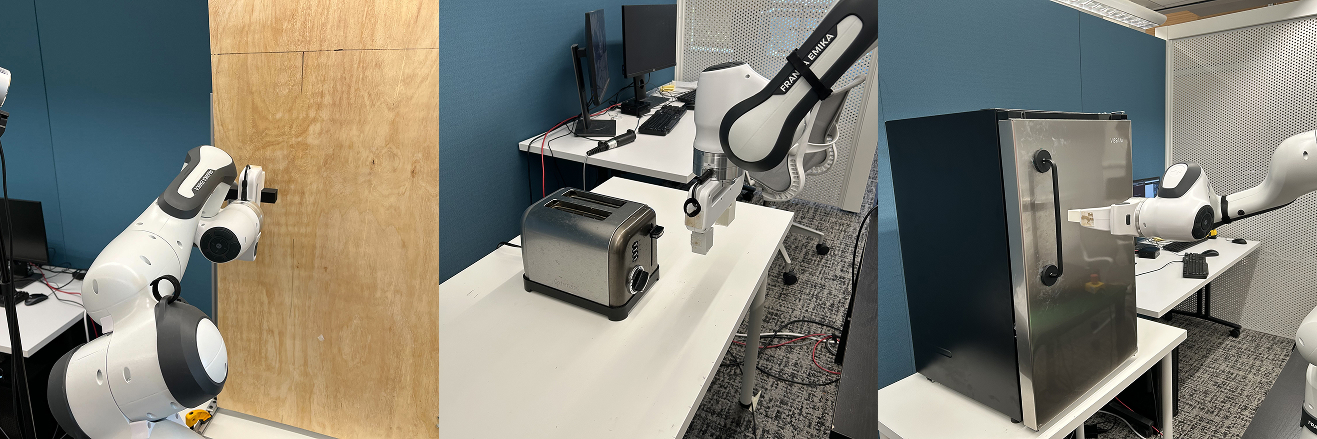}
    \caption{Real world environment setup. From left to right: push open door, push toaster lever, and pull open fridge door.}
    \label{fig:real_envs}
\end{figure}

We also show that Infinigen-Articulated assets are effective in sim-to-real transfer. We create real-world versions of the door opening, toaster pushing, and fridge opening tasks using a Franka Panda robot (see \figref{fig:real_envs}). For each task, we collect expert simulation trajectories using cuRobo \cite{curobo_report23} in an IsaacGym \cite{makoviychuk2021isaacgym} environment. We randomize environment lighting, joint initialization, object position, and camera views. As we have the ground truth joint axis and origin, we roll out scripted trajectories and filter successful ones as the demonstration dataset. The robot takes an RGB image from its on-the-shoulder camera and the end effector pose as inputs. We then train an ACT \cite{zhao2023aloha} policy on this dataset and transfer it zero-shot to a physical robot. 

\begin{table}[H]
\centering
\renewcommand{\arraystretch}{1.2}
\begin{tabular}{l|ccc}
      & Open Door  & Push Toaster & Pull Fridge \\ \hline
PartNet-Mobility \cite{Xiang_2020_SAPIEN}    &  3/30  &  0/30  &  0/30 \\
Infinigen-Articulated (Ours) &  \textbf{18/30}  &  \textbf{11/30}  &  \textbf{10/30}
\end{tabular}
\caption{Zero-shot sim-to-real results.}
\label{tab:real_results}
\end{table}

We observe that policies trained using Infinigen-Articulated perform significantly  better across all three tasks (see \tableref{tab:real_results}). Qualitatively, we notice that policies trained using our assets tend to be smoother and fail less aggressively. For instance, failures in the fridge task by the Infinigen-Articulated policy typically involved either moving toward the handle and nearly grasping it or opening the door less than the required 45 degree threshold. In contrast, policies trained on PartNet-Mobility often missed the handle altogether. Similar to the RL experiments, we attribute this success to the increased diversity and quantity of assets generated with our tool. We postulate that it is this diversity that helps bridge the sim-to-real gap. 

\section{Limitations and Future Work}

Joints are currently modeled independently. While coupling them (e.g., pressing the pedal of a trash can to open the lid) is a relatively straightforward extension, we leave this for future work. Additionally, Infinigen-Articulated currently uses CoACD \cite{wei2022coacd} to generate collision meshes for individual parts. Depending on the level of detail of the convex decomposition, this can potentially slow simulation. That being said, a key advantage of Infinigen-Articulated is that we know how each part is constructed via its node graph. We leave it as future work to determine how to generate a corresponding set of primitives to serve as the asset's collision mesh using the node graph, potentially improving simulation speed.

\section{Conclusion}

We introduce Infinigen-Articulated, a toolkit to generate procedural, articulated, simulation-ready assets. We use the system to provide procedural generators for 18 articulated object categories. These assets are kinematically diverse, photorealistic, follow consistent naming conventions, and have accurate joint configurations. Infinigen-Articulated provides users fine-grained control over their assets including sampling from physical parameters to bridge the sim-to-real gap. Our experiments demonstrate that objects from these generators can improve model generalization on both perception and robot learning tasks in simulation and the real world.

\section*{ACKNOWLEDGMENT}
We thank Lingjie Mei for helping with modifications to the door asset. This work was partially supported by a Cisco grant and the National Science Foundation.

\bibliographystyle{bibtex/IEEEtran}
\bibliography{bibtex/IEEEabrv,bibtex/IEEEexample}

\section*{APPENDIX}

\subsection{Articulated Parts}
\label{sec:articulated_parts}
\tableref{tab:articulated_objects} provides a list of articulated parts for each of our procedural generators. Certain parts may have their own procedural parameters including style, count, positioning, dimensions, etc.

\begin{table}[h]
\centering
\begin{tabular}{l l}
\toprule
\textbf{Object Category} & \textbf{Articulated Parts} \\
\midrule
Door          & Door, handle \\
Cooktop       & Burner dial \\
Lamp          & Lamp arms, shade, switch/button/dial/pull chain \\
Box           & Box flaps \\
Soap dispenser& Nozzle \\
Oven          & Door, racks, drawer \\
Toaster       & Lever, buttons, knobs \\
Cabinet       & Doors \\
Drawer        & Doors \\
Pepper grinder& Top bulb/handle \\
Refrigerator  & Doors, internal drawers, external drawers \\
Trashcan      & Lid, pedal \\
Dishwasher    & Door, buttons, racks, knobs \\
Faucet        & Spout, handles \\
Microwave     & Door, internal plate, eject button, dials \\
Window        & Window panes, locks \\
Plier         & Plier hand \\
Handle        & Main handle, lock \\
\bottomrule
\end{tabular}
\caption{Provided object categories and their articulations.}
\label{tab:articulated_objects}
\end{table}

\subsection{Moveable Part Segmentation Experiment Details}

To curate a dataset, we generate images using segmentation ground truth for both the PartNet-Mobility and Infinigen-Articulated assets. Each image is of size $1024 \times 1024$ to ensure that smaller articulated parts are visible. Similar to \cite{Xiang_2020_SAPIEN}, images are taken from a camera positioned randomly on the upper hemisphere of the asset with the scene having ambient lighting and fixed directional lighting. We also randomly sample the field of view. Finally, we randomize joint values such that the same asset appears in different poses across images. For assets such as dishwashers where interior racks may intersect with exterior bodies, we sample joint values within a range to try to avoid this situation. All images are rendered using ray-tracing with Maniskill3 \cite{taomaniskill3}.

We then randomly sample PartNet-Mobility assets for our evaluation set such that there exists at least one asset for each considered movable part. To ensure fairness, we remove invalid assets from the evaluation set based on the following criteria:

\begin{itemize}[leftmargin=0.2in, topsep=0pt, itemsep=0pt, parsep=0pt]
    \item Asset includes parts that should be movable but are not.
    \item Asset contains invalid geometries such as holes or protruding triangles.
    \item Asset contains incorrect face normals.
\end{itemize}

We subsample the evaluation dataset such that we have 25\% of assets from each model category, ensuring at least one asset remains for each movable part. Remaining assets are moved back to the PartNet-Mobility training dataset. Hyperparameters for training are shown in \tableref{tab:maskrcnn_hyperparameters}.

\begin{table}[h]
\centering
\begin{tabular}{l c}
\toprule
\textbf{Hyperparameter} & \textbf{Value} \\
\midrule
Learning Rate           & $1 \times10^{-4}$ \\
Weight Decay            & $1 \times10^{-5}$ \\
Evaluation Mask Threshold & 0.5 \\
Batch Size              & 4 \\
Optimizer               & Adam \\
\bottomrule
\end{tabular}
\caption{Mask R-CNN training hyperparameters.}
\label{tab:maskrcnn_hyperparameters}
\end{table}

\subsection{RL Generalization Experiment Details}

We define our reward functions as shown below where $p_x$ is the position of the $x$, $\hat q_{x}$ is the normalized joint configuration, $\lambda_a = 0.1$ is the action penalty scale, and $T(x) = 1 - \tanh (x)$. If the task is successful, we assign an $R_{\text{success}}$ reward greater than the maximum reward defined by the functions below.
{\small
\begin{align*}
    r_{\text{door}}(s,a) &= T\!\left(10 \,\lVert p_{\text{eef}} - p_{\text{handle}} \rVert \right) 
                      + 4T\!\left(3(1-\hat{q}_{\text{handle}}) \right) \\
                      &\quad + 5T\!\left(2 (1 - \hat{q}_{\text{door}}) \right) 
        - \lambda_a \lVert a \rVert \\
    r_{\text{toaster}}(s,a) &= T\!\left(10 \,\lVert p_{\text{eef}} - p_{\text{lever}} \rVert \right) 
                      + 2T\!\left(3(1-\hat{q}_{\text{lever}}) \right) - \lambda_a \lVert a \rVert \\
    r_{\text{fridge}}(s,a) &= T\!\left(5 \,\lVert p_{\text{eef}} - p_{\text{handle}} \rVert \right) 
                      + 2(1 - (1 - \hat q_{\text{door}})^4) - \lambda_a \lVert a \rVert 
\end{align*}
}

Hyperparameters for all RL experiments are shown in \tableref{tab:rl_hyperparameters}.

\begin{table}[h]
\centering
\begin{tabular}{l c}
\toprule
\textbf{Hyperparameter} & \textbf{Value} \\
\midrule
Learning Rate           & $1 \times10^{-4}$ \\
Gamma                   & 0.95 \\
Number of Parallel Environments  & 128 \\
\bottomrule
\end{tabular}
\caption{PPO training hyperparameters.}
\label{tab:rl_hyperparameters}
\end{table}

\subsection{Sim-to-Real Transfer Experiment Details}

We generate expert trajectories for each task using a Franka Panda robot in IsaacGym \cite{makoviychuk2021isaacgym}. During each step, we store both the robot's fingertip's pose and an RGB image from an on-the-shoulder, third-person camera (rendered using IsaacGym's default renderer). To generate trajectories, we sample candidate pre-grasps and grasp poses according to the handle position for the door and fridge tasks and the lever position for the toaster task. An operation space controller (OSC) is then used to generate a trajectory. We design specific paths for each task:  

\begin{itemize}
    \item \textbf{Pushing open doors:} linear motion to the handle, circular motion to rotate the handle 45 degrees (based on handle–joint distance), and circular motion to push the door open (based on handle–hinge distance).
    \item \textbf{Pushing toaster lever:} linear motion to the lever.  
    \item \textbf{Pulling open fridge:} linear motion to the handle, grasp, and circular motion to pull the door open (based on handle–hinge distance).
\end{itemize}

We train our policy on fingertip poses and RGB images. The policy outputs a target 6-DOF pose for the arm in the robot base frame and a single value for the gripper action instead of delta motion \cite{zhao2023aloha}. The training configuration follows \cite{zhao2023aloha}. We train multiple policies with different hyperparameters for 300 epochs and select the checkpoint with the lowest validation loss. The final hyperparameters are described in \tableref{tab:act_hyperparameters}. We use Deoxys \cite{zhu2022viola} to command the physical robot with an OSC controller for the absolute target pose. We set a relative stiff PD value.

\begin{table}[h]
\centering
\begin{tabular}{l l c c c}
\toprule
\textbf{Dataset} & \textbf{Hyperparameter} & \textbf{Door} & \textbf{Toaster} & \textbf{Fridge} \\
\midrule
\multirow{3}{*}{Ours} 
  & Learning Rate        & $1 \times10^{-4}$  & $5 \times10^{-5}$ & $1 \times10^{-4}$  \\
  & KL Weight            & 1  & 10  & 100  \\
  & Temp. Agg. &  \ding{55}  & $\checkmark$  & $\checkmark$   \\
\midrule
\multirow{3}{*}{PartNet} 
  & Learning Rate        & $1 \times10^{-4}$  & $1 \times10^{-4}$  & $5 \times10^{-5}$  \\
  & KL Weight            & 10  & 10  & 10  \\
  & Temp. Agg. & \ding{55}   & \ding{55}  & $\checkmark$   \\
\bottomrule
\end{tabular}
\caption{Final ACT policy hyperparameters for each task on Infinigen-Articulated and PartNet-Mobility.}
\label{tab:act_hyperparameters}
\end{table}

\subsection{Comment on Robot Safety}

For the real-world fridge opening task, the handle occasionally slipped into a gap between the robot’s fingers, causing it to get stuck. When the robot pulled back, this created excessive torque on the gripper and raised safety concerns. To address this, we add padding between the door and the base of the fridge as well as on the gripper to close the gap.

\end{document}